\documentclass[acmmedium, review=false]{acmart}
\AtBeginDocument{%
  }

\renewcommand\footnotetextcopyrightpermission[1]{} 
\usepackage{setspace}
\makeatletter
\renewcommand\normalsize{%
  \@setfontsize\normalsize{9pt}{10pt}%
}
\makeatother
\normalsize

\acmJournal{POMACS}
\acmVolume{37}
\acmNumber{4}
\acmArticle{111}
\acmMonth{8}

\begin{document}

\title{Can LLMs Capture Expert Uncertainty? A Comparative Analysis of Value Alignment in Ethnographic Qualitative Research}


\author{Arina Kostina}
\email{akosti02@ucy.ac.cy}
\affiliation{%
  \institution{University of Cyprus}
  \city{Nicosia}
  \country{Cyprus}
}

\author{Marios Dikaiakos}
\email{mdd@cs.ucy.ac.cy}
\affiliation{%
  \institution{University of Cyprus}
  \city{Nicosia}
  \country{Cyprus}
}

\author{Alejandro Porcel}
\email{ap2251@cam.ac.uk}
\affiliation{%
  \institution{Trinetra Investment Management LLP}
  \city{London}
  \country{UK}
}
\affiliation{%
  \institution{University of Cambridge}
  \city{London}
  \country{UK}
}

\author{Tassos Stassopoulos}
\email{tassos.stassopoulos@trinetra-im.com}
\affiliation{%
  \institution{Trinetra Investment Management LLP}
  \city{London}
  \country{UK}
}

\renewcommand{\shortauthors}{Kostina et al.}

\begin{abstract}

Qualitative analysis of open-ended interviews plays a central role in ethnographic and economic research by uncovering individuals’ values, motivations, and culturally embedded financial behaviors. While large language models (LLMs) offer promising support for automating and enriching such interpretive work, their ability to produce nuanced, reliable interpretations under inherent task ambiguity remains unclear. In our work we evaluate LLMs on the task of identifying the top three human values expressed in long-form interviews based on the Schwartz Theory of Basic Values framework. We compare their outputs to expert annotations, analyzing both performance and uncertainty patterns relative to the experts. Results show that LLMs approach the human ceiling on set-based metrics (F1, Jaccard) but struggle to recover exact value rankings, as reflected in lower RBO scores. While the average Schwartz value distributions of most models closely match those of human analysts, their uncertainty structures across the Schwartz values diverge from expert uncertainty patterns. Among the evaluated models, Qwen performs closest to expert-level agreement and exhibits the strongest alignment with expert Schwartz value distributions. LLM ensemble methods yield consistent gains across metrics, with Majority Vote and Borda Count performing best. Notably, systematic overemphasis on certain Schwartz values, like Security, suggests both the potential of LLMs to provide complementary perspectives and the need to further investigate model-induced value biases. Overall, our findings highlight both the promise and the limitations of LLMs as collaborators in inherently ambiguous qualitative value analysis.

\end{abstract}





\maketitle

\begingroup
\renewcommand\thefootnote{}
\footnotetext{This submission has been accepted for a poster session at BIG.AI@MIT 2026.}
\endgroup

\section{Introduction}

Qualitative analysis of open-ended interviews constitutes a key component of ethnographic research, enabling the systematic identification and interpretation of individuals’ values, lived experiences, and latent motivations~\cite{Ritchie:2013}. In recent years, such qualitative approaches have gained increasing prominence within economics~\cite{Skarbek:JInstEcon:2020:QualitativeRM, Chopra:AEA:2024}, where they offer rich, interpretable context about the motivations behind consumption behaviors. Understanding the subtle, culturally influenced dynamics of financial habits helps investors identify opportunities emerging from shifts in the values, needs, and priorities of consumers. Large language models (LLMs) offer a new dimension to qualitative research, with the potential of not only automating tedious tasks but also assisting in more creative aspects, like offering new perspectives in data interpretation. The availability of LLMs through natural-language interfaces substantially reduces the entry threshold for researchers without technical training. Some previous work has indicated that LLMs struggle to accurately capture language structure and subtle meanings \cite{Korngiebel:NPJDigMed:2021, vanDijk:ArXiv:2023}. However, manual interpretation of long, unstructured interview data is both time-consuming and inherently subjective, relying heavily on the perspectives and judgments of individual researchers. As increasingly capable models emerge at a rapid pace, systematic and ongoing evaluation of their performance on complex linguistic tasks is essential to determine the extent to which LLMs can support qualitative researchers in producing nuanced and reliable analyses.

Recent LLM progress points to a promising direction for interpretive work. Beyond automating labor-intensive stages of qualitative analysis, these systems can also assist in more creative tasks, like offering new perspectives in data interpretation. However, LLM integration into such workflows introduces concerns about interpretive robustness, particularly in high-stakes domains such as investment research: while LLM are known to generate responses that are fluent and convincing, it is often unclear whether these outputs reflect grounded reasoning or stem from model bias and hallucination.

Collaborative ethnographic analysis typically involves multiple researchers engaging in critical discussion, bringing diverse perspectives to bear on the same material. Given that human experts themselves are shaped by their environments, education, and cultural backgrounds, disagreements and opposing opinions can occur. Crucially, these disagreements are not merely inevitable but productive, serving as a vital mechanism for surfacing and counteracting the personal biases inherent in individual expert judgments. Importantly, uncertainty in qualitative value analysis is also intrinsic to the task itself, as human values are often overlapping, context-dependent, implicit, and only partially articulated by interviewees. Participants may express motivations indirectly, and sensitive or complex financial topics can further obscure clear value attribution. While much of the existing LLM evaluation literature prioritizes aggregate performance metrics, frequently overlooking the inherent complexity and ambiguity of qualitative interpretation, a critical issue remains underexplored: \textit{Can LLMs produce interpretations and uncertainty patterns comparable to those of human analysts under inherent task ambiguity?} 



In this work we conduct a case study based on 12 open-ended, 2-hour interview sessions with local residents in China, using LLMs to identify the dominant value orientations expressed by participants, as defined by the Schwartz Theory of Basic Human Values. Unlike most research relying on self-reported questionnaires, we infer values contextually, posing a challenge for LLMs to interpret implicit human behavior and sociocultural meaning without explicit keywords. Our objectives mainly focus on: 

\begin{enumerate}
    \item Conducting a value alignment analysis, addressing the questions of whether LLMs align with expert-derived value distributions and whether their uncertainty reflects expert disagreement. Specifically, we examine whether intra-model variability across alternative prompt formulations mirrors inter-expert variability, potentially indicating sensitivity to values that are inherently ambiguous, context-dependent, or difficult to infer.
    
    \item Performing a case-study on the use of LLMs on a real-world long and unstructured set of interviews in an interpretive task, and to evaluate whether aggregating outputs from multiple models via an LLM ensemble improves overall value identification performance. 
    
    \item Exploring performance and consistency of individual LLMs, and influence of different prompting and input segmentation strategies. 
    
\end{enumerate}



\section{Related Work}

Qualitative analysis and anthropology have gained significant traction in investing as a means of uncovering decision-making processes underlying latent trends in consumer behavior~\cite{Starr:2014:QualitativeAM,Skarbek:JInstEcon:2020:QualitativeRM,Stassopoulos:2022,Twilley:Wired:2024}. Although qualitative analysis processes can provide insights of unparalleled depth and contextual grounding, their design and execution are inherently time-intensive, costly, and susceptible to human error, especially when handling large corpora, as they depend heavily on the sustained involvement of trained human experts. Often experts have to come back to the interview, verify and cross-reference conclusions. This underscores the potential benefit of computational approaches that can help accelerate the analysis process without compromising depth, accuracy, and nuance. 

The adoption of Qualitative Data Analysis Software (QDAS) has facilitated content analysis by improving data organization and retrieval. However, these tools primarily support data management and still require researchers to interpret findings and make analytic decisions manually ~\cite{Woods:SocSciCompRev:2016, Rasheed:ArXiv:2024:LLMAsDataAnalyst}. More recently, the emerging field of computational ethnography has explored the use of machine learning (ML) for the automation and augmentation of ethnographic analysis tasks; for example, using unsupervised  learning to perform topic modeling ~\cite{Owler:Frontiers:2023:MLQualitative}, supervised classification and natural language processing (NLP) to assist concept coding in large qualitative datasets ~\cite{Rietz:CHI:2021}, and helping researchers dialogically interrogate their qualitative data~\cite{McKay:PublicHealth:2025}. However, these models often require large, carefully curated, and heavily preprocessed datasets to achieve reasonable performance in complex or nuanced research questions, and such datasets are typically scarce in qualitative analysis~\cite{Xiao:ACM:2023:SupportingQA}. 

In recent years, especially since the launch of OpenAI’s ChatGPT \cite{brown:ArXiv:2020:language, openai:2023:gpt4}, LLMs have gained prominence as powerful AI-driven models excelling in various NLP tasks, including text summarization, classification, translation, and question-answering. Their success in other domains has sparked extensive research into their potential applications in qualitative analysis. LLM-assisted qualitative coding~\cite{Xiao:ACM:2023:SupportingQA, Kirsten:ArXiv:2024:decodingcomplexityexploringhumanai} demonstrate that careful codebook adaptation, and task decomposition can yield human-comparable performance, while multi-agent framework can accelerate the qualitative data analysis process \cite{Rasheed:ArXiv:2024:LLMAsDataAnalyst}. Although LLMs generate compelling content that appears factual and implies a capacity for reasoning, they do not possess genuine understanding or an inherent notion of communicative intent or truth, due to their probabilistic nature ~\cite{Vaswani:NeurIPS:2017:AttentionIA}, leading to hallucinations in their outputs ~\cite{Bender:FAccT:2021:StochasticParrots,Ji:ACM:2023:HallucinationSurvey}. In addition, LLMs may generate distorted output, which reflects biases and stereotypes embedded in their training corpora ~\cite{jaffe:Microsoft:2024:productivity,dikaiakos:politikon:2025,paschalides:ICWSM:2025}. A line of prior work highlights bias risks in LLM-based coding \cite{Ashwin:arxiv:2023:UsingLLMs}, and proposes systematic value-mapping frameworks like FULCRA to validate Schwartz mappings and detect model-specific inconsistencies \cite{Yao:ArXiv:2023:valuefulcramappinglarge}. Whereas much of the existing literature concentrates on identifying potential biases in LLM-generated responses, our work instead evaluates model performance relative to human experts, emphasizing a realistic analytical setting where even expert judgments diverge and vary due to the task’s intrinsic ambiguity. Most prior studies on values analyze datasets consisting of short text snippets, typically around one paragraph, often curated to highlight information indicative of specific values, while we focus on long-format open-ended interviews, which contain substantial portions of text that refer indirectly to value orientation, presenting a challenge for LLMs in identifying meaningful cues. 

\section{Methodology}
\subsection{Data, Annotation, and Metrics}
Our dataset comprises transcripts from 12 in-person interviews conducted with participants from China, who were selected and interviewed according to protocols developed and applied by an organization undertaking ethnographic studies worldwide. The interview questions focused on people's 
hopes, anxieties, aspirations and the social issues that their families face. The interviews were conducted in Chinese, recorded and automatically transcribed, and manually translated by expert translators to English, with the expert translation also being recorded and transcribed. Each interview followed an unstructured question-and-answer format. A multi-disciplinary team of six experts in ethnographic analysis (anthropologists, economists, investment specialists) independently annotated each interview with the most prominent nuanced Schwartz values~\cite{Schwartz1994} (from the set of 58 values), which were then mapped to their corresponding 10 basic motivational types defined by the Schwartz theory of human values. Final ground truth was determined via majority voting across annotators, producing the top 3 values per interview. The Krippendorff’s $\alpha$ across annotators for this task was 0.389, which indicates the inherent ambiguity in value assignment task. To help contextualize model performance under the computed task complexity we compute a human-ceiling baseline via leave-one-annotator-out evaluation, with results reported in Table \ref{tab:schwartz_best_model}. This method measures how well individual experts align with the aggregated judgments of the other experts. The metrics include F1-score (F1@3), Jaccard similarity (Jaccard@3), which quantifies set intersection, and Rank-Biased Overlap (RBO@3), which captures top‐weighted overlap in indefinite rankings. For the LLM evaluation, all our metrics are calculated for the first 3 values identified as most important by the LLMs, since our ground truth consists of 3 values for each interview. 
%

\subsection{LLM models}
We evaluated state-of-the-art, open-source decoder-only models with a minimum context window of 32,000 tokens, to ensure they are able to process the entire interview in a single prompt. The models were obtained from the HuggingFace platform, from the Unsloth provider, and run via llama.cpp library in GGUF-quantized format with quantization size chosen based on the availability for the specific model, and hardware constrains. Specifically, we assessed (1) DeepSeek-R1-Distill-Llama-8B with Q4\_K\_M precision \cite{DeepSeekAI:2025:DeepSeekR1IR}, (2) Qwen3-30B-A3B-Instruct-2507 with Q8\_0 precision \cite{qwen3}, (3) Llama-3.3-70B-Instruct with Q5\_K\_M precision \cite{llama3:meta:2024}, and Mistral-Small-3.2-24B-Instruct-2506 with Q8\_0 precision \cite{mistral:small24b}. We used the models’ recommended temperature, changing the seed when the model produced degenerate generation, as was particularly observed with the DeepSeek and Mistral models. The models were hosted on HPC servers, with four Tesla V100 GPUs (32GB each). 

\subsection{Prompt Engineering}
Prompt engineering involves designing instructions to guide LLMs for specific tasks. We iteratively refined prompts by testing them first on a sample transcript, and then applying them across all interviews. Prompts were evaluated in two settings: using the whole transcript, and segments of around 5,000 tokens each, ensuring splits did not break sentences or words. For segmented inputs, an aggregation prompt combined outputs of individual responses for each segment into a final response. Prompts include the following techniques and combinations of them: (1) \textbf{Baseline} prompt which asks for the prioritization of the 10 Schwartz values, along with the explanations; (2) \textbf{Bias-Constraint Prompting (BC)}, which specifies \textit{"Maintain
complete objectivity—there are no good or bad values."}, aiming to mitigate biases in LLM outputs, as certain values may be implicitly treated as negative and under-assigned, while others may be over-assigned due to their generally positive connotations in the training data \cite{Zhao:2025:ArXiv:Explicitvsimplicitinvestigating}; (3) \textbf{Profile-Enhanced Prompting (PEP)}, which includes a short summary of the interviewee’s background in the prompt, thereby offering contextual cues, which helps LLMs reproduce those characteristics of the persona \cite{Miotto:ArXiv:2022:gpt3explorationpersonalityvalues}; and (4) \textbf{Bottom-Up Prompt (BUP)}, which aims to imitate the bottom-up annotation process of the experts - starting from 58 subvalues, and aggregating into a final ranking of the 10 broad values.

\section{Result Analysis}

\subsection{Performance Evaluation and Prompting Influence}

Table \ref{tab:schwartz_best_model} demonstrates substantial performance variability across prompting techniques and segmentation strategies. For each model, the reported mean represents the average performance across the eight prompt-segmentation configurations, where each configuration is first averaged over the performance on each interview. The corresponding standard deviation reflects the variability across the prompt configurations, quantifying prompt sensitivity. The results show that models exhibit substantial variability across prompting and segmentation strategies, especially DeepSeek, with its mean scores being low and standard deviations being equal to or larger than the mean. Llama, Mistral, and Qwen3 achieve higher average performance, but still show standard deviations of 20–27 points. In most cases, this variability exceeds that of the human-ceiling by around 5 points, with the exception of RBO, where expert disagreement is higher. Qwen approaches the human-ceiling the closest, with average F1 of 56.6 and Jaccard of 43.96, but has a notably lower RBO score of 37.09. Llama and Mistral perform slightly lower but remain in a comparable range for the F1 and Jaccard metrics. Overall, the results suggest that identifying the correct top three values under non-ordered metrics (F1 and Jaccard) is comparatively easier than accurately capturing their ranking, as reflected by the consistently lower RBO scores.

For prompting and segmentation influence, Table \ref{tab:schwartz_best_prompt} illustrates that Profile-Enhanced Prompting (PEP) in whole interview setting achieves the highest scores across most metrics, while the Bottom-Up approach achieved inferior performance in both segmentation settings across most metrics. It is also notable that the influence of granularity of the input varies, as PEP and BUP perform better with the whole transcript, while the Baseline and BC + PEP perform slightly better when the interviews are split.

Finally, intra-model reliability results, calculated with Krippendorff’s $\alpha$ show moderate agreement across different prompt phrasings. Qwen and Llama achieve the highest intra-model agreement, with $\alpha$ values of 0.548 and 0.540 respectively. Mistral exhibits lower but still moderate agreement at 0.393, while DeepSeek shows very low consistency across prompts, with an $\alpha$ of 0.128.

\subsection{Performance of LLM Ensemble Method}

We utilize the LLM Ensemble approach, which is based on combining the responses of multiple LLMs \cite{Chen:ArXiv:2025:harnessingmultiplelargelanguage}. 
Each LLM is individually prompted with the same prompting and segmentation technique, producing its own prioritization of the values. The values are then aggregated with one of the following rank aggregation methods: (1) Kemeny-Young \cite{Hemaspaandra:2005:KemenyYoungVoting}, (2) Majority Vote, and (3) Borda Count \cite{Saari:1985:BordaCount}. We use a leave-one-model-out analysis, evaluating ensembles formed from all combinations of three out of four models. This also maintains an odd number of voters for rank aggregation, reducing the likelihood of ties across the answers. Table \ref{tab:aggregation_comparison} reports the $\Delta$ of the aggregation strategies, meaning the performance improvement achieved by the LLM Ensemble compared to the average of the individual (standalone) models. Results indicate that all aggregation methods improve performance, achieving a performance gain of 8-10 points on F1 and RBO metrics, and 6-8 points on Jaccard similarity. Majority Vote and Borda Count show the strongest and nearly identical gains, with Majority vote achieving slightly higher Jaccard similarity.

\subsection{Distribution \& Uncertainty Analysis in LLM Value Predictions}

In real expert analysis, higher disagreement may indicate that some values are inherently ambiguous in the provided interview context, or difficult to infer. If an LLM shows higher variability for the same values that experts find ambiguous, this suggests the LLM may be sensitive to underlying uncertainty of the values and material, while a mismatch would attribute the uncertainty to model-specific factors, which has important implications for trust, interpretation, and downstream decision-making. To explore these concerns, we examine the global, and interview-level distribution of Schwartz values across our interview corpus. Our variability analyses measures intra-model agreement, meaning for each model we compute variability across different prompting strategies, quantifying the variability of responses within each model. In this setup, each prompt acts as a distinct interpretive lens, allowing us to assess how consistently a single model assigns values under alternative formulations of the same task. 

At the interview level, we first examine alignment between the mean per-value distributions of experts and the models. For this we compute cosine similarity on the mean per-value distribution of each model (across prompts) and mean per-value distribution of experts. Second, we examine whether model uncertainty mirrors expert disagreement by calculating Spearman’s $rho$ between the per-value standard deviation of model predictions (across prompts) and the per-value standard deviation of expert annotations. This evaluates whether the model is uncertain about the same values that humans find ambiguous. Additionally, we report median per-value standard deviation across Schwartz values, which captures the overall magnitude of a model's uncertainty across Schwartz values. This metric complements Spearman's $rho$, which measures only whether the model's uncertainty pattern aligns with that of experts, remaining insensitive to absolute variability. A model could achieve high $rho$ while being systematically over- or under-confident. Together, the two metrics distinguish between misalignment in uncertainty structure and misalignment in overall uncertainty magnitude. 
For all metrics, we report bootstrap means and 95\% confidence intervals at the interview level, using resampling to account for variability across interviews and obtain more reliable estimates of our metrics. 

At the global level, we examine value distributions across the entire interview corpus by summing assignments for each value over all interviews. For each model, we compute the mean and standard deviation across prompts, and for experts, across expert annotations. The global distributions complement the interview-level analysis, helping identify potential systematic over- or under-representation of specific values. We plot the results visually to facilitate comparison.

Table \ref{tab:variance_alignment} reveals differences across models in how accurately they approximate the mean value distributions of the experts' assignments. Qwen shows the closest alignment to the experts' value distribution in the responses ($cosine=0.833$), with relatively tight confidence intervals. Mistral and Llama follow closely with comparable performance of $cosine=0.817$ and $cosine=0.795$ respectively. In contrast, DeepSeek exhibits markedly lower mean alignment of $0.552$, with wider confidence intervals than other models. While all models achieve relatively high cosine similarities (>0.79, except DeepSeek), Table \ref{tab:variance_alignment} shows that models demonstrate a weaker alignment with expert variability structure across the values. Qwen demonstrates the closest value uncertainty to the human experts, though the correlation is still moderate ($rho=0.457$). Mistral falls behind by a substantial margin ($rho=0.379$), followed by Llama and lastly Deepseek.  
Table \ref{tab:variance_bootstrapping} measures the median per-value standard deviation, which indicates how much a model's output varies across prompts. Notably,  Llama is the most consistent model ($std = 0.147$), followed by Qwen3 ($std = 0.202$), both exhibiting lower variability across prompts than inter-expert variability, which suggests systematic overconfidence as the cause of value uncertainty misalignment ($rho$).
In contrast, DeepSeek exhibits uncertainty magnitudes almost perfectly aligned with expert disagreement ($std = 0.254$), yet differs substantially in the mean value distributions ($cosine$), and uncertainty pattern ($rho$) across values. Mistral ($std = 0.280$) displays the highest internal variability across the prompts. 

\begin{figure*}[t]
    \centering
    \includegraphics[width=0.9\textwidth]{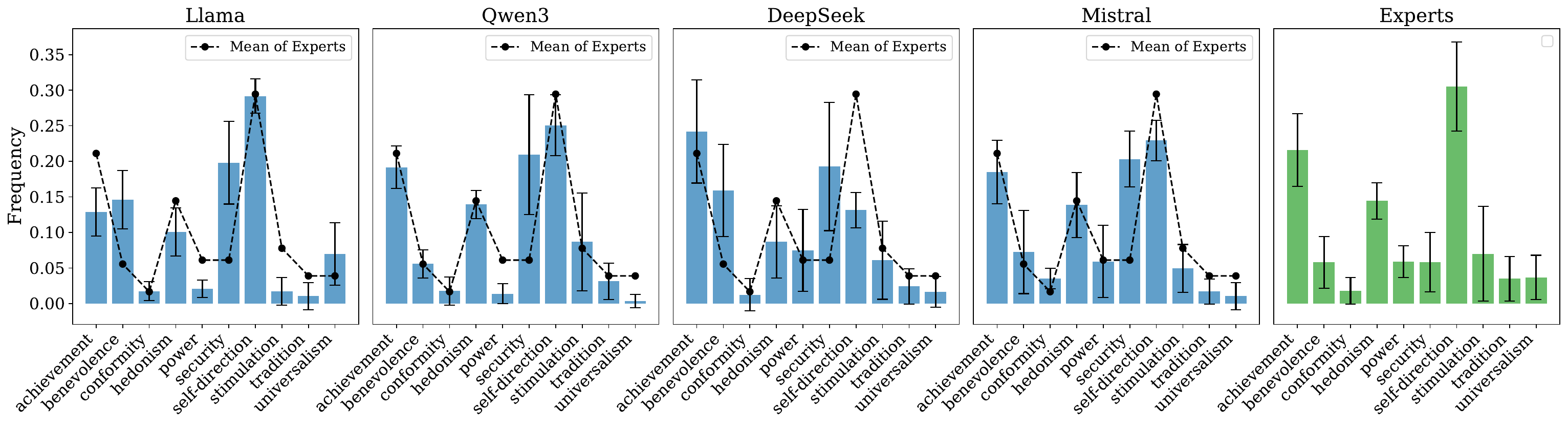}
    \caption[] %
{\small Global Level Distribution Analysis. Schwartz value distributions across all interview corpus, with prompting variance for each LLM, are compared to the experts’ distribution. } 
\label{fig:schwartz_bias}
\end{figure*}

Table \ref{fig:schwartz_bias} presents the global value distributions of the models in comparison to the experts. The observed discrepancies are largely value-specific and differ across models. However, interestingly, all the models seem to assign the \textit{Security} value a lot more frequently than the experts. This discrepancy can reveal noteworthy patterns and potentially overlooked insights in the data, suggesting that model outputs may provide a complementary viewpoint. At the same time, prior work shows that training data and methodology can induce value biases in LLMs, raising questions about whether their value assignments reflect grounded reasoning or are influenced by other factors \cite{Benkler:ArXiv:2023:Assessingllmsmoralvalue}. 



\section{Conclusion, Limitations, and Future Work}

Overall, our findings show that LLMs achieved a strong performance in the task of top-3 value identification, approaching the human ceiling in the set-based metrics (F1 and Jaccard), while struggling more with recovering the exact ranking of values, as reflected in lower RBO scores. Among the models, Qwen consistently performs closest to the human ceiling and shows the strongest alignment with expert value distributions and uncertainty patterns, exhibiting relatively low variance across prompts. At the same time, performance is sensitive to prompting and input segmentation, with variability of models across configurations often exceeding expert variability and indicating moderate intra-model reliability. All models appear to have a close average value distribution with the experts, while their uncertainty structure across values tends to diverge from expert uncertainty patterns. Input segmentation showed varying performance across prompts, while the Profile-Enhanced prompting with the whole interview setting achieved the best results. LLM ensemble gives consistent performance gains of around 8-10 points on F1 and RBO, and 6-8 on Jaccard, over standalone models across all aggregation strategies. More specifically, Majority Vote and Borda Count proved most effective. Finally, systematic discrepancies in value assignment, notably the overemphasis on Security, underscore the potential of LLMs to provide complementary insights, but also prompt us to investigate potential model-induced value biases in future work. With our work we provide insights into the capabilities and limitations of LLMs in the interpretive work within qualitative research workflows. While the primary aim of the paper was to provide qualitative insights and an exploratory analysis, a larger set of interview samples would be beneficial for producing more generalizable results and statistical claims. Ongoing work includes evaluating an additional preprocessing step that utilizes LLMs to clean the interview transcripts and restructure them into numbered question–answer pairs, which allows referencing of supporting text for the value assignments. Future work will extend the evaluation to a broader set of models, including closed-source LLMs. 

\begin{table}
\begin{minipage}{0.49\textwidth}
    \centering
    \begin{tabular}{ll|ccc}
    \toprule
    & & \multicolumn{3}{|c}{\textbf{Average Results}} \\
    \toprule
    \textbf{Prompt} & \textbf{Inp} & \textbf{F1@3}  & \textbf{Jaccard@3} & \textbf{RBO@3} \\
    \toprule
    BUP & W & 46.11 & 33.33 & 29.05 \\
    Baseline & W & 46.67 & 36.35 & 30.03 \\
    BC + PEP & W & 44.72 & 33.75 & 29.69 \\
    PEP & W & 49.03 & 38.85 & 32.29 \\
    \midrule
    BUP & S & 41.32 & 30.07 & 25.23  \\
    Baseline & S & 47.50 & 35.83 & 32.64  \\
    BC + PEP & S & 46.67 & 35.52 & 31.02 \\
    PEP & S & 45.76 & 34.24 & 30.21  \\
    \midrule
\end{tabular}
\caption{\small Schwartz's Value Theory - Average scores per prompt, aggregated across all models for Whole (W) and Split (S) settings}
\label{tab:schwartz_best_prompt}
    \label{tab:first}
\end{minipage}
\hfill
\begin{minipage}{0.49\textwidth}
\renewcommand{\arraystretch}{0.8}  
\caption{Average scores per model, aggregated across all prompts, with the standard deviation, and the human ceiling scores.
}
\label{tab:schwartz_best_model}

    \centering
    
    \begin{tabular}{lccc}
    \toprule
    \textbf{Model} & \textbf{F1@3} & \textbf{Jaccard@3} & \textbf{RBO@3} \\
    \toprule
    DeepSeek & 24.90 ± 25.02 & 16.75 ± 18.08 & 17.88 ± 21.33 \\
    Llama & 52.08 ± 26.85 & 40.00 ± 27.38 & 34.72 ± 23.75 \\
    Mistral & 50.31 ± 27.24 & 38.26 ± 26.58 & 30.38 ± 21.02 \\
    Qwen3 & 56.60 ± 25.65 & 43.96 ± 26.14 & 37.09 ± 20.48 \\
    
    \midrule

    Experts & 58.19 ± 22.63 & 44.54 ± 22.48 & 51.97 ± 32.31 \\
    \midrule



\end{tabular}

\centering
\begin{tabular}{lccc}

\toprule
\textbf{Method} & \textbf{F1@3} & \textbf{Jaccard@3} & \textbf{RBO@3} \\
\toprule
Kemeny-Young & $8.20 \pm 1.74$ & $5.81 \pm 1.93$ & $7.83 \pm 2.75$ \\
Majority Vote & $9.51 \pm 2.51$ & $8.00 \pm 2.62$ & $10.30 \pm 2.92$ \\
Borda Count & $9.51 \pm 2.40$ & $7.08 \pm 2.50$ & $10.69 \pm 2.51$ \\
\midrule
\end{tabular}
\caption{\small Comparison of $\Delta$ of aggregation strategies in LLM Ensemble across evaluation metrics (mean ± std) ($\Delta$ = Ensemble performance - Average standalone model performance).}
\label{tab:aggregation_comparison}

\end{minipage}

\end{table}

\begin{table}
\begin{minipage}{0.45\textwidth}
\centering
\small
\begin{tabular}{lccc}
\toprule
Model & Median $std$ & CI$_{low}$ & CI$_{high}$ \\
\midrule
Llama & 0.147 & 0.089 & 0.205 \\
Qwen3 & 0.202 & 0.133 & 0.270 \\
DeepSeek & 0.254 & 0.199 & 0.303 \\
Mistral & 0.280 & 0.232 & 0.314 \\
\midrule
Experts & 0.252 & 0.155 & 0.338 \\
\bottomrule
\end{tabular}
\caption{\small Bootstrap means and 95\% confidence intervals for median per-value standard deviation across prompts (for models) and across experts, with interview-level bootstrapping.}
\label{tab:variance_bootstrapping}
\end{minipage}
\hfill
\begin{minipage}{0.54\textwidth}
\centering
\small
\begin{tabular}{lccc|ccc}
\toprule
Model & Mean $\rho$ & CI$_{low}$ & CI$_{high}$   & Mean Cosine & CI$_{low}$ & CI$_{high}$ \\
\midrule
Llama & 0.328 & 0.257 & 0.400    & 0.795 & 0.732 & 0.856  \\
Qwen3 & 0.457 & 0.345 & 0.579    & 0.833 & 0.793 & 0.868  \\
DeepSeek & 0.309 & 0.135 & 0.469    & 0.552 & 0.454 & 0.645  \\
Mistral & 0.379 & 0.220 & 0.532    & 0.817 & 0.741 & 0.879  \\
\bottomrule
\end{tabular}
\caption{\small Bootstrap means and 95\% confidence intervals for variance alignment (Spearman's $\rho$) and mean alignment (cosine similarity). Metrics were computed between model and expert mean for cosine, std for $\rho$, value distributions, averaged across prompts, with interview-level bootstrapping.}
\label{tab:variance_alignment}
\end{minipage}
\end{table}

\bibliographystyle{ACM-Reference-Format}
\bibliography{sample-base}

\appendix

\end{document}